\documentclass[10pt,twocolumn,letterpaper]{article}

\usepackage[pagenumbers]{iccv}

\definecolor{iccvblue}{rgb}{0.21,0.49,0.74}
\usepackage[pagebackref,breaklinks,colorlinks,allcolors=iccvblue]{hyperref}

\usepackage{pgfplots}
\pgfplotsset{compat=1.18}

\usepackage{xspace}

\usepackage{algorithm}
\usepackage{amsmath}
\usepackage{algpseudocode}  % For \Comment and advanced formatting
\usepackage{etoolbox}       % For \hrulefill spacing adjustment
\usepackage{multirow}
\usepackage{tabularx}
\usepackage[accsupp]{axessibility}

\title{Aligning Vision to Language: Annotation-Free Multimodal Knowledge Graph Construction for Enhanced LLMs Reasoning}

\author{Junming Liu$^{1,2}$, Siyuan Meng$^{2,3}$, Yanting Gao$^{1}$, Song Mao$^{2}$, Pinlong Cai$^{2}$,\\
Guohang Yan$^{2}$, Yirong Chen$^{2,4}$, Zilin Bian$^{5}$, Ding Wang$^{2}$\thanks{Corresponding author.}, Botian Shi$^{2}$\\
$^{1}$Tongji University\quad
$^{2}$Shanghai Artificial Intelligence Laboratory\\
$^{3}$East China Normal University\quad
$^{4}$Stanford University\quad
$^{5}$New York University\\
{\tt\small liu\_junming6917@tongji.edu.cn\quad
wangding@pjlab.org.cn
}}
\begin{document}
\maketitle

\begin{abstract}
Multimodal reasoning in Large Language Models (LLMs) struggles with incomplete knowledge and hallucination artifacts, challenges that textual Knowledge Graphs (KGs) only partially mitigate due to their modality isolation. While Multimodal Knowledge Graphs (MMKGs) promise enhanced cross-modal understanding, their practical construction is impeded by semantic narrowness of manual text annotations and inherent noise in visual-semantic entity linkages.
In this paper, we propose \textbf{V}ision-align-to-\textbf{L}anguage \textbf{i}ntegrated \textbf{K}nowledge Graph (VaLiK), a novel approach
for constructing MMKGs that enhances LLMs reasoning through cross-modal information supplementation.
Specifically, we cascade pre-trained Vision-Language Models (VLMs) to align image features with text, transforming them into descriptions that encapsulate image-specific information.
Furthermore, we developed a cross-modal similarity verification mechanism to quantify semantic consistency, effectively filtering out noise introduced during feature alignment.
Even without manually annotated image captions, the refined descriptions alone suffice to construct the MMKG.
Compared to conventional MMKGs construction paradigms, our approach achieves substantial storage efficiency gains while maintaining direct entity-to-image linkage capability.
Experimental results on \textit{multimodal reasoning tasks} demonstrate that LLMs augmented with VaLiK outperform previous state-of-the-art models.
Our code is published at \href{https://github.com/Wings-Of-Disaster/VaLiK}{https://github.com/Wings-Of-Disaster/VaLiK}.
\end{abstract}    
\section{Introduction}
\label{sec:intro}

Recent advancements in Large Language Models (LLMs)~\cite{Chen_2020_GPT3, Touvron_2023_Llama2, Achiam_2023_Gpt4, Guo_2025_Deepseekr1} have demonstrated their superiority and versatility across various Natural Language Reasoning (NLR) tasks~\cite{Robinson_2022_Leveraging, Chen_2024_Benchmarking, Plaat_2024_Reasoning, Liu_2025_Mosaic}. To enhance LLMs into the realm of multimodal reasoning, researchers~\cite{Yin_2023_Survey, Wu_2024_Semantic, Xu_2024_Mlevlm, Tong_2024_Cambrian1} have endeavored to equip these models with multimodal capabilities, as evidenced by advancements in Multimodal Large Language Models (MLLMs) such as BLIP-2~\cite{Li_2023_BLIP2}, GPT-4o~\cite{Hurst_2024_Gpt4o}, Janus-Pro~\cite{Chen_2025_Janus}, among others. Despite their notable progress, these models often experience \textit{hallucinations}~\cite{Azamfirei_2023_Large, Kalai_2024_Calibrated}, primarily arising from knowledge deficiencies due to incomplete or obsolete information.

\begin{figure}[t]
	\centering
	\includegraphics[width=1.0\columnwidth]{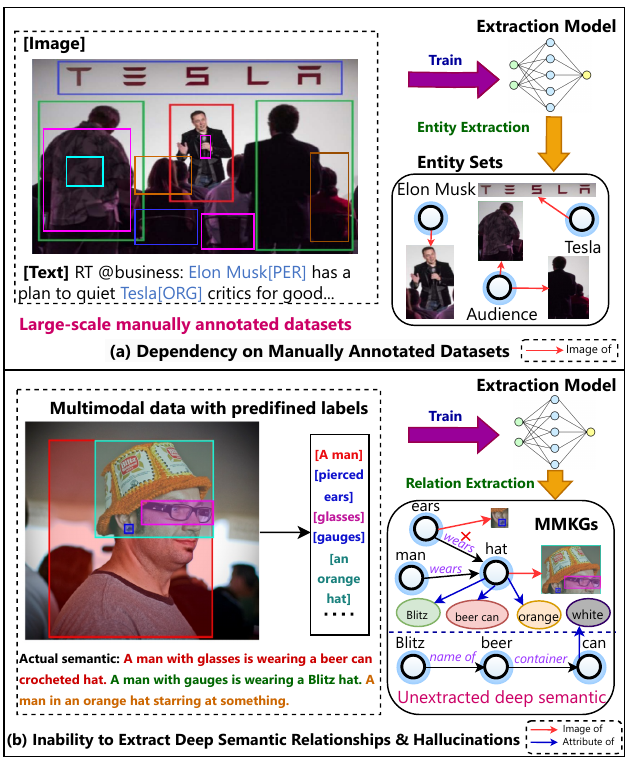}
	\caption{(a) Training entity extraction models relies on extensive fine-grained annotations, increasing labeling costs. More examples are provided in Appendix B. (b) Capturing implicit semantic associations demands abstract comprehension or logical inference.}
        \label{fig:motivation}
    \vspace{-4mm}
\end{figure}

Fine-tuning LLMs demands prohibitive computational costs~\cite{Hu_2023_LLM}. While text-based Knowledge Graphs (KGs) have partially addressed this limitation by efficient real-time updates~\cite{Sun_2021_Ernie, Wu_2023_Retrieve, Baek_2023_Knowledge}, they are still restricted by modal isolation, which hinders cross-modal reasoning, as detailed in Appendix A. To bridge this semantic fragmentation, Multimodal Knowledge Graphs (MMKGs) have been developed as unified representational frameworks~\cite{Johnson_2015_Image, Liu_2019_MMKG, Chen_2022_Hybrid, Lee_2024_Multimodal}.

However, constructing robust MMKGs faces two primary obstacles~\cite{Chen_2024_Knowledge, Zhu_2024_Multimodal}. First, the lack of large-scale fine-grained entity-image corpora makes it infeasible to train high-quality entity extractors, significantly constraining scalability, as illustrated in Figure~\ref{fig:motivation}a.
Second, conventional visual relation detectors primarily identify superficial spatial interactions instead of semantic relations consistent with KGs, while frequently hallucinating implausible connections that corrupt graph integrity, as shown in Figure~\ref{fig:motivation}b.

\begin{figure}[t]
	\centering
	\includegraphics[width=1.0\columnwidth]{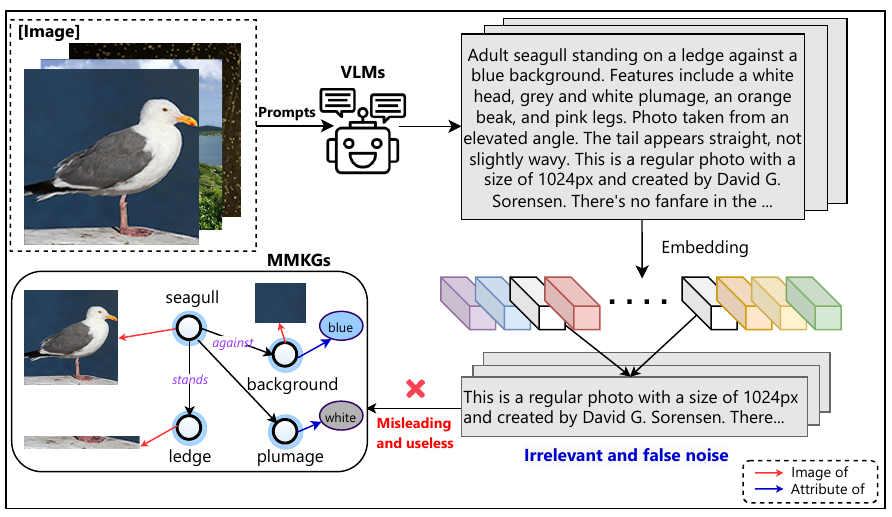}
	\caption{Feature-aligned descriptions from VLMs introduce redundant and inaccurate relationship patterns.}
        \label{fig:phenomenon}
    \vspace{-4mm}
\end{figure}

In this paper, we propose VaLiK, short for Vision-align-to-Language integrated Knowledge Graph, a novel framework designed to empower LLMs with advanced multimodal reasoning. Unlike traditional methods that rely on text annotations for training extraction models and the knowledge construction process~\cite{Plummer_2015_Flicker30k}, VaLiK adopts a annotation-free approach to MMKGs construction.
Specifically, we first employ several pretrained Vision-Language models (VLMs), designed based on Chain-of-Experts (CoE) principles~\cite{Xiao_2024_CoE}, to convert visual inputs into image-specific textual descriptions through cross-modal feature alignment. This procedure eliminates the need for manually annotated image captions in both the knowledge extraction and construction phases while preserving visual details typically missing in generic text descriptions.
Moreover, in contrast to existing relation detection methods that require predefined label taxonomies~\cite{Zheng_2021_Multimodal, Cui_2024_Enhancing, Song_2024_Scene, Zhang_2024_MyGo}, VaLiK excels at extracting profound semantic relationships that are both KG-compatible and capture novel associations beyond training supervision. While VLMs enable cross-modal reasoning and interpretation, they introduce spurious relational noise through hallucinated inter-modal attributions, as depicted in Figure~\ref{fig:phenomenon}. We address this limitation through cross-modal similarity recalibration, strategically filtering inconsistent information while preserving valid semantic correspondences.
Finally, the purified descriptions are systematically organized into MMKGs via LLM-driven symbolic structuring~\cite{Guo_2024_LightRAG}, bridging visual and textual domains with factual consistency.

To thoroughly evaluate the VaLiK method, we conduct a comprehensive assessment across two critical multimodal benchmarks: \textit{multimodal classification} (tested on the CrisisMMD dataset~\cite{Alam_2018_CrisisMMD}) and \textit{multimodal question answering} (evaluated via the ScienceQA benchmark~\cite{Lu_2022_ScienceQA}). The experiments span diverse LLM architectures and MMKG construction techniques to ensure the framework’s robustness. The experimental results demonstrate that the MMKGs constructed by VaLiK achieve superior multimodal reasoning performance in LLMs while requiring substantially less storage than conventional approaches. More importantly, the proposed approach retains direct entity-to-image linkage capabilities even with the compressed graph structure. 

In summary, VaLiK is the first framework that enables \textit{end-to-end}, \textit{annotation-free}, \textit{zero-shot}, and \textit{storage-efficient} multimodal knowledge construction with high adaptability and scalability. Our key contributions include:
\begin{itemize}[noitemsep,nolistsep]
\item To the best of our knowledge, VaLik is the first end-to-end framework to build Annotation-Free MMKGs to improve LLMs' multimodal reasoning capabilities, effectively eliminating the need for manually annotated textual material and enabling a completely autonomous multimodal knowledge generation process.
\item We offer an innovative zero-shot method for constructing MMKG that captures deep semantic connections beyond traditional predetermined labels with an effective verification system that guarantees the accuracy of these relationships. The knowledge distillation paradigm greatly decreases storage while maintaining semantic integrity.
\item We develop a highly modular and extensible architecture that allows VaLiK to effortlessly incorporate new models and workflows for specialized domain tasks, facilitating rapid adaptation to diverse application scenarios without incurring expensive system changes.
\end{itemize}

\section{Related Work}
\label{sec:RL}

\subsection{Multimodal Knowledge Graphs}

The principal advantage of MMKGs resides in their multimodal integration beyond conventional KGs. By linking entities with corresponding visual or textual data, MMKGs introduce valuable visual and textual information to the knowledge base, substantially advancing multimodal reasoning capabilities.
This combination addresses core challenges in tasks that inherently demand multimodal synergy like autonomous driving~\cite{Guo_2023_Visual, Halilaj_2023_KGADD}, image-text retrieval~\cite{Feng_2023_MKVSE, Zheng_2024_KGITR} and robotic manipulation~\cite{Reily_2020_Representing, Miao_2023_Semantic}.
However, constructing trustworthy MMKGs with minimal manual effort remains a critical challenge. Recent studies have proposed innovative strategies to enhance MMKG reliability and utility. For instance, Chen \etal\cite{Chen_2024_Continual} proposed MSPT, a framework addressing continual MMKG construction through gradient modulation for balanced multimodal learning and attention distillation to mitigate catastrophic forgetting. Song \etal\cite{Song_2024_Scene} developed Scene-MMKG, integrating knowledge engineering with large language models to improve robotic manipulation by resolving data sparsity and knowledge uncertainty. Wang \etal\cite{Wang_2023_TIVAKG} introduced TIVA-KG, the first quad-modal knowledge graph spanning text, image, video, and audio with triplet grounding, empirically validating its effectiveness in downstream tasks. While these advances enhance multimodal reasoning capabilities, their efficacy remains rooted in resource-intensive paradigms, requiring extensively annotated datasets for knowledge acquisition. 

\subsection{Knowledge-Augmented Multimodal Learning}

Multimodal learning has seen significant progress in aligning and integrating information across different data modalities~\cite{Baltrušaitis_2019_Multimodal, Xu_2023_Multimodal, Liu_2025_FedRecon}.
The incorporation of structured knowledge through MMKGs further enhances these approaches, improving the reasoning capabilities and generalization across a variety of domains, such as visual question answering~\cite{Wang_2018_FVQA, Marino_2019_OKVQA, Singh_2021_MIMOQA}, recommendation systems~\cite{Sun_2020_Multimodal, Wang_2020_Enhanced, Cui_2023_MKGCN}, and classification~\cite{Zhao_2020_Knowledge, Qian_2021_Knowledge, Hu_2022_Graph}. Methods like GraphAdapter’s dual-KG adaptation~\cite{Li_2023_GraphAdapter} and contrastive multi-relational encoding with KGs~\cite{Fang_2023_Contrastive} inject external knowledge into models, refining their performance and improving their capability to handle complex tasks. Additionally, Lee \etal\cite{Lee_2024_Multimodal} proposed MR-MKG, a novel framework that constructs task-specific MMKGs to enhance multimodal reasoning in LLMs. These knowledge-augmented paradigms demonstrate superior cross-modal semantic grounding compared to unimodal approaches~\cite{Kannan_2020_Multimodal, Chen_2023_Suevey}.
However, their reliance on preconstructed MMKGs often leads to domain discrepancies, where generic knowledge schemas misalign with task-specific reasoning patterns, ultimately limiting contextual precision in target applications.

\subsection{Multimodal Large Language Models}

The limitations of text-only LLMs in meeting increasingly complex demands have spurred extensive research~\cite{Ye_2023_Mplug, Zheng_2023_Advances, Zhang_2024_Multimodal} into developing LLMs capable of effectively processing and reasoning over multimodal inputs. Current research predominantly employs adapter or projection layers to connect the embedding spaces of various modality-specific encoders with the textual embedding space of LLMs~\cite{Lee_2024_Multimodal}. For instance, foundational models like CLIP~\cite{Radford_2021_Learning} and BLIP~\cite{Li_2022_Blip} pioneered cross-modal alignment by jointly training vision and text encoders to map images and text into a shared embedding space. Building on this, LLaVA~\cite{Liu_2023_LLaVa} and Flamingo~\cite{Alayrac_2022_Flamingo} advanced the field by integrating visual encoders with LLMs, enabling more nuanced multimodal understanding and generation. More recently, Gemini~\cite{Google_2023_Gemini}, Qwen2-VL~\cite{Wang_2024_QWEN2VL} and GPT-4o~\cite{Hurst_2024_Gpt4o} have further pushed the boundaries by scaling up multimodal pretraining and introducing sophisticated mechanisms for cross-modal interaction.
However, multimodal LLMs remain prone to hallucinations. While they enhance cross-modal alignment, they neither acquire new knowledge nor avoid introducing noise through integration. To address these limitations, VaLiK "uses the master's tools to refine the master's craft," first constructing MMKGs via MLLMs and then leveraging them to enhance MLLMs' reasoning capabilities.
\section{Method}

\begin{figure*}[t]
	\centering
	\includegraphics[width=1.0\textwidth]{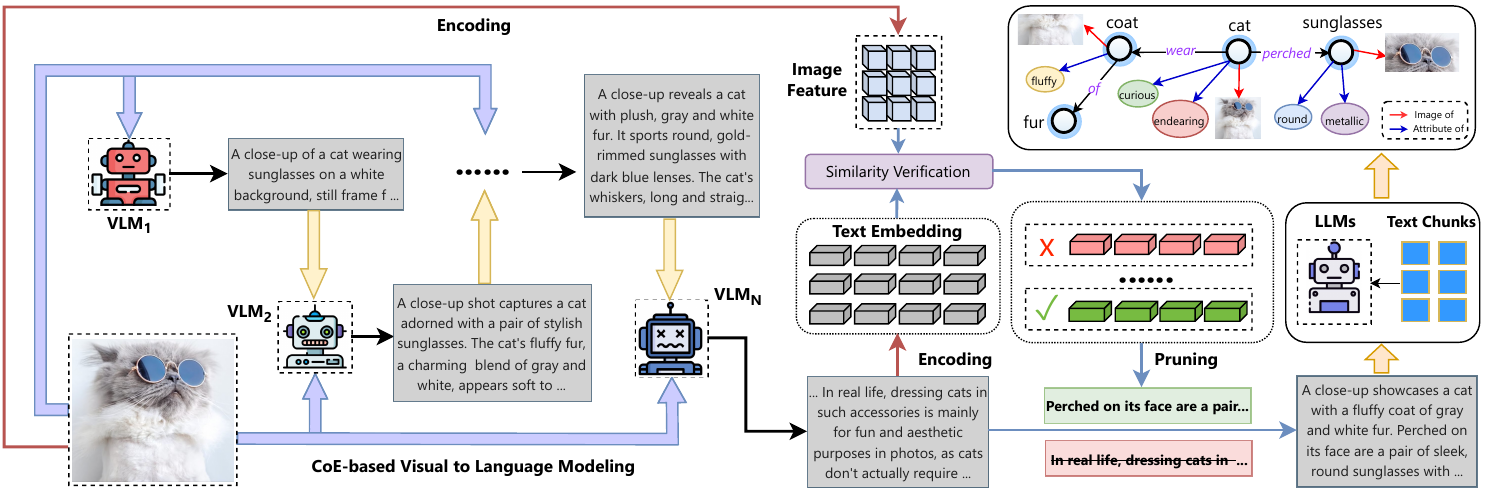}
	\caption{The pipeline of VaLiK: First, large-scale visual descriptions are generated using CoE-based VLMs. Then, a similarity verification mechanism is used to prune irrelevant information. Finally, MMKGs are constructed using LLMs based on LightRAG. The constructed MMKGs can assist LLMs in multimodal reasoning, alleviating the hallucination issues caused by incomplete knowledge.}
    \label{fig:pipeline}
    \vspace{-4mm}
\end{figure*}

In this section, we present the technical details of VaLiK. VaLiK introduces a novel expansion-reduction paradigm for visual knowledge extraction.
The architecture initially organizes several VLMs with distinct knowledge domains, designed based on CoE principles~\cite{Xiao_2024_CoE}, to produce comprehensive textual descriptions encompassing hierarchical visual details.
A cross-modal similarity verification mechanism then iteratively filters out noisy tokens through cross-modal alignment while preserving semantically salient elements. This optimization-style approach eliminates external textual dependencies while enabling effective MMKG construction. VaLiK's framework is shown in Figure~\ref{fig:pipeline}.

\subsection{CoE-based Visual to Language Modeling}
\label{subsec:v2t }

Recent entity detection techniques~\cite{Diwan_2022_Object, Zou_2023_Object, Yue_2025_Roburcdet} have been widely adopted for entity and relation extraction in MMKG construction.
However, these methods are inherently limited by predefined categorical boundaries, lacking the capacity to recognize visual concepts outside their training vocabulary. In contrast, VLMs pretrained on web-scale corpora~\cite{Li_2023_BLIP2, Chen_2023_Pali, Zhu_2024_Minigpt4} exhibit broader recognition capabilities through exposure to diverse visual concepts.

We therefore leverage pretrained VLMs to extract comprehensive visual information. This process removes the necessity for detailed fine-grained data typically required to train specialized recognition models. The generalized vision to language conversion pipeline can be formalized as:
\begin{equation}
S = \mathcal{D}_{\text{text}}\Big( \mathcal{A}\big( \mathcal{E}_{\text{vis}}(I) \big) \Big),
\end{equation}
where $I$ denotes for the input image, $\mathcal{E}_{\text{vis}}$ denotes the visual encoder extracting visual features, $\mathcal{A}$ carries out cross-modal feature alignment and interaction, and $\mathcal{D}_{\text{text}}$ generates textual tokens through autoregressive decoding. The resulting visual description $S = \{w_1, ..., w_n\}$ emerges from this multi-stage processing.

However, quantitative analysis uncovers considerable discrepancies between machine-generated and human-annotated descriptions~\cite{Zhu_2023_Chatcaptioner}. As an illustration, while utilizing BLIP-2~\cite{Li_2023_BLIP2} to generate sample captions, we noted that the model outputs are markedly concise and devoid of visual specifics, as detailed in Appendix C. To bridge this gap, we implement CoE enhanced generation through cascade VLMs processing. At iteration step $t$, each expert $E_i$ receives both the original visual signals $I$ and the contextual output from the preceding expert $E_{i-1}$:
\begin{equation}
\mathcal{S}^{(t)}_i = E_i\left(I, \mathcal{S}^{(t-1)}_{i-1}\right),
\end{equation}
where $\mathcal{S}^{(t-1)}_{i-1}$ denotes the description from expert $E_{i-1}$ at step $t-1$, with $\mathcal{S}^{(t)}_0 := \varnothing$ for initialization.

Specifically, each expert $E_i$ implements a unified visual-language processing task:
\begin{enumerate}
    \vspace{0.2em}
    \item \textbf{Visual Feature Extraction:}
    \begin{equation}
        \mathbf{V}_i = \text{Enc}_{\text{vis}}^i(I) \in \mathbb{R}^{d_v \times N_p},
    \end{equation}
    where $\text{Enc}_{\text{vis}}^i$ denotes established visual encoder~\cite{He_2016_ResNet, Dosovitskiy_2020_ViT, Liu_2021_Swin} producing $N_p$ patch embeddings with dimension $d_v$.
    
    \vspace{0.2em}
    \item \textbf{Cross-Modal Interaction and Generation:}
    
    VLMs integrate pretrained learnable query embeddings $\mathbf{Q}_i \in \mathbb{R}^{d_q \times L_q}$ to interact with visual features $\mathbf{V}_i \in \mathbb{R}^{d_v \times N_p}$ via cross-attention~\cite{Vaswani_2017_Attention}:
    \begin{equation}
    \begin{aligned}
    \mathbf{H}_i &= \text{CrossAttn}(\mathbf{Q}_i, \mathbf{V}_i) \\
    &= \text{softmax}\left(\frac{\mathbf{Q}_i\mathbf{W}_q^i (\mathbf{V}_i\mathbf{W}_k^i)^\top}{\sqrt{d_k}}\right)\mathbf{V}_i\mathbf{W}_v^i,
    \end{aligned}
    \end{equation}
    where $\mathbf{W}_q^i \in \mathbb{R}^{d_q \times d_k}$, $\mathbf{W}_k^i, \mathbf{W}_v^i \in \mathbb{R}^{d_v \times d_k}$, and $L_q$ denotes the predefined query length.
    Cross-attention serves as a prevalent approach, while other interaction strategies coexist~\cite{Alayrac_2022_Flamingo}. The adopted VLMs in our implementation primarily rely on this approach for modality fusion.

    \vspace{0.2em}
    \item \textbf{Text Generation:}
    
    The text encoder $\text{Enc}_{\text{text}}^{i}$ first processes the preceding expert's output $\mathcal{S}^{(t-1)}_{i-1}$ into latent features: 
    \begin{equation}
    \mathbf{P}_{i} = \text{Enc}_{\text{text}}^{i}(\mathcal{S}^{(t-1)}_{i-1}) \in \mathbb{R}^{d_t \times L}.
    \end{equation}
    Subsequently, the text decoder $\text{Dec}_{\text{text}}^i$ synthesizes the final output $\mathcal{S}^{(t)}_i$ by jointly conditioning on $\mathbf{P}_i$ and $\mathbf{H}_i$:  
    \begin{equation}
    \mathcal{S}^{(t)}_i = \text{Dec}_{\text{text}}^i(\mathbf{P}_i, \mathbf{H}_i) = \{w_1^{(t,i)}, \ldots, w_m^{(t,i)}\}.
    \end{equation}  
\end{enumerate}
Ultimately, the final textual description $\mathcal{S}^{(C)}_{N}$ is obtained after $C$ iteration steps through $N$ cascaded experts.

\subsection{Cross-Modal Similarity Verification}
\label{subsec:denoise}

To address noise in VLM-generated captions, we design a sliding window mechanism with semantic consistency verification. This method ensures that only relevant and semantically consistent segments are retained in the final description. Let $W_k$ denote the $k$-th window containing $m$ consecutive tokens $\{w_{km+1}, ..., w_{(k+1)m}\}$. For each window, we compute its cross-modal similarity score:
\begin{equation}
    \alpha_k = \frac{\text{Enc}_{\text{vis}}(I) \cdot \text{Enc}_{\text{text}}(W_k)}{\|\text{Enc}_{\text{vis}}(I)\| \|\text{Enc}_{\text{text}}(W_k)\|},
\end{equation}
where $\text{Enc}_{vis/text}(\cdot)$ adopts a lightweight CLIP~\cite{Robinson_2022_Leveraging} encoder-decoder with frozen parameters for efficient processing.
The similarity score \( \alpha_k \) lies within the range [0, 1], with higher values indicating a stronger alignment between the visual and textual information. 

After calculating the cross-modal similarity for each window, we employ an empirical threshold \( \tau \) to filter out low-similarity windows. This threshold helps to identify and discard noisy or irrelevant sections of the generated caption that do not align well with the visual content, thereby reducing the impact of inaccurate or misleading descriptions. Formally, for each window \( W_k \), if \( \alpha_k < \tau \), the window is discarded as noise. This process effectively \textbf{prunes} windows with low similarity scores, ensuring that only semantically meaningful segments remain. The final denoised description \( \hat{S} \) is obtained by concatenating all windows \( W_k \) for which \( \alpha_k \geq \tau \):
\begin{equation}
    \hat{S} = \bigcup_{\alpha_k \geq \tau} W_k.
\end{equation}
Our window size \(m\) is flexibly determined and generally adapts dynamically to natural sentence segmentation.

\subsection{MMKG Construction for Enhanced Reasoning}
\label{subsec:kg}

LLMs have become increasingly popular for identifying entities, relationships, and attributes within a corpus, which are then organized into a KG. The strength of LLM-based KG generation lies in its capacity to leverage the vast amount of knowledge encoded within these models, allowing them to detect complex and nuanced patterns across diverse data sources. This approach eliminates the need for manual annotation, enabling a highly scalable and domain-adaptive process suitable for a wide range of applications.

We begin by refining the generated textual description \( \hat{S} \) (VLM-based information), which is then optionally concatenated with any available external textual knowledge \( T \) to form the input for KG generation.
This combined input is used to generate MMKGs with the help of a LLM~\cite{Edge_2024_GraphRAG, Guo_2024_LightRAG}, leveraging its capacity for multi-hop reasoning and dynamic knowledge integration.

\begin{equation}
    \mathcal{G} = \text{LLM}\left(\hat{S} \oplus T\right),
\end{equation}
where \(\oplus\) denotes optional concatenation based on the availability of \(T\). The resulting graph \(\mathcal{G}\) captures both visual and textual relationships inferred by the LLM.

We define \(\mathcal{G}\) as a set of triplets:
\begin{equation}
    \mathcal{G} = \{(h,r,t) \mid h,t \in \mathcal{E}, r \in \mathcal{R}\},
\end{equation}
where \(\mathcal{E}\) and \(\mathcal{R}\) denote the sets of entities and relations.
Entities include objects or concepts from the image or external text, while relations describe connections such as “is a type of,” “part of,” or “has property.” Each triplet \((h, r, t)\) links a head entity \(h\) and a tail entity \(t\) via relation \(r\).

\noindent\textbf{Multimodal Reasoning Enhancement.}
To support multimodal reasoning, we retrieve relevant triplets from \( \mathcal{G} \) through structural patterns during LLMs inference:
\begin{equation}
    \mathcal{G}_q = \text{Retrieve}(q, \mathcal{G}),
\end{equation}
where \(\text{Retrieve}(\cdot)\) denotes a retrieval strategy that identifies subgraphs relevant to the query for reasoning. Detailed retrieval strategies are described in Appendix D.

The augmented prompt integrates multimodal evidence:
\begin{equation}
    p_{\text{aug}} = q \parallel \big( \bigcup_{(h,r,t) \in \mathcal{G}_q} [h]{\rightarrow}r{\rightarrow}[t] \big).
\end{equation}
Note that we incorporate the storage locations of images in the database during MMKGs construction, enabling the MMKGs to link to visual data. VaLiK enables text-only LLMs to perform multimodal reasoning through $\mathcal{G}$'s visual associations, while VLMs refresh knowledge representations by jointly injecting both visual and textual information, significantly mitigating hallucination risks.
\section{Experiment}

\begin{table*}[t!]
    \centering
    \footnotesize
    \scalebox{0.92}{
    \begin{tabular}{l *{15}{c}}
        \toprule
        \multirow{2}{*}{\textbf{Task}} & 
        \multicolumn{11}{c}{\textbf{Text-only LLMs}} & 
        \multicolumn{3}{c}{\textbf{KG-Enhanced LLMs}} \\
        \cmidrule(lr){2-12} \cmidrule(lr){13-15}
        & 
        \multicolumn{3}{c}{LLaMA-2} & 
        \multicolumn{1}{c}{GPT-4} &
        \multicolumn{4}{c}{DeepSeek-R1} &
        \multicolumn{3}{c}{Qwen2.5} & 
        \multicolumn{1}{c}{LightRAG} & 
        \multicolumn{2}{c}{VaLiK} \\
        \cmidrule(lr){2-4} \cmidrule(lr){5-5} \cmidrule(lr){6-9} \cmidrule(lr){10-12}
        \cmidrule(lr){13-13} \cmidrule(lr){14-15}
        & 7B & 13B & 70B & - & 7B & 8B & 32B & 70B & 7B & 32B & 72B & Text-only & Image-only & Text-Image \\
        \midrule
        Task 1 & 62.32 & 63.80 & 63.15 & 66.83 & 67.23 & 63.31 & 63.61 & 65.53 & 65.04 & 67.28 & 67.95 & 67.49 & \textbf{69.52} & \underline{68.90} \\
        Task 2 & 18.32 & 21.82 & 28.87 & 47.25 & 26.53 & 25.49 & 24.77 & 21.05 & 44.52 & 46.94 & \textbf{50.51} & 45.11 & 49.54 & \underline{50.02} \\
        Task 2 Merged & 21.45 & 33.15 & 36.89 & 49.44 & 25.85 & 23.56 & 21.55 & 25.57 & 45.33 & 47.07 & \underline{50.29} & 45.94 & 49.07 & \textbf{50.69}\\
        \bottomrule
    \end{tabular}
    }
    \caption{The performance evaluation of text-only LLMs using few-shot prompting without any fine-tuning on the training set.
    As these models handle text only, test data is formatted as unimodal text for compatibility. In our implementations, both LightRAG and VaLiK adopt \textbf{Qwen2.5-7B} as the base reasoning model.
    \textbf{Bold} indicates the highest value, and \underline{underline} indicates the second highest.
    }
    \label{tab:text_kg}
\end{table*}

\begin{table*}[t!]
    \newcommand{\sym}[2][0]{\raisebox{0.3ex}{\scalebox{1.3}{\raisebox{#1ex}{#2}}}}
    \centering
    \footnotesize
    \scalebox{1.00}{
    \begin{tabular}{l *{14}{c}}
        \toprule
        \multirow{2}{*}{\textbf{Task}} & 
        \multicolumn{10}{c}{\textbf{Multimodal VLMs}} & 
        \multicolumn{4}{c}{\textbf{KG-Enhanced LLMs}} \\
        \cmidrule(lr){2-11} \cmidrule(l){12-15}
        & 
        \multicolumn{1}{c}{CLIP} & 
        \multicolumn{3}{c}{LLaVA} & 
        \multicolumn{2}{c}{BLIP-2} &
        \multicolumn{1}{c}{GPT-4o} &
        \multicolumn{3}{c}{Qwen2-VL} & 
        \multicolumn{4}{c}{VaLiK} \\
        \cmidrule(lr){2-2} \cmidrule(lr){3-5} \cmidrule(lr){6-7}
        \cmidrule(lr){8-8} \cmidrule(lr){9-11} \cmidrule(l){12-15}
        & ViT-L/14 & 7B & 13B & 34B & Flan-T5-XL & OPT & - & 2B-I & 7B-I & 72B-I & \sym[-0.5]{*} & \sym[-0.2]{\#} & \sym[-0.2]{+} & \sym[-1.0]{\textasciitilde} \\ 
        \midrule
        Task 1 & 43.36 & 54.00 & 60.58 & 56.44 & 61.29 & 38.62 & 68.20 & 47.56 & 62.45 & 65.80 & 60.78 & \underline{68.44} & 61.11 & \textbf{68.89} \\
        Task 2 & 17.88 & 28.01 & 20.14 & 25.15 & 40.86 & 14.26 & 47.58 & 7.60 & 32.68 & 47.21 & 25.80 & \underline{48.88} & 27.23 & \textbf{49.78} \\
        Task 2-M & 20.79 & 30.61 & 23.44 & 25.07 & 40.72 & 14.27 & \textbf{49.55} & 7.42 & 34.20 & 48.28 & 27.31 & 49.27 & 29.09 & \underline{49.31} \\
        \bottomrule
    \end{tabular}
    }
    \caption{The performance of multimodal VLMs and KG-enhanced LLMs. The -I suffix denotes instruction-tuned variants. Symbol markers denote KG types and models: the asterisk (*) represents image-only KG with LLaVA-34B, hash (\#) indicates image-only KG using Qwen2-VL-72B-I, plus (+) denotes text-image KG with LLaVA-34B, and tilde (\textasciitilde) shows text-image KG using Qwen2-VL-72B-I.}
    \label{tab:multi_kg}
\end{table*}

\subsection{Setups}

\textbf{Evaluation Datasets.} We evaluate VaLiK on two multimodal reasoning benchmarks with distinct characteristics:
\begin{itemize}[noitemsep,nolistsep]
\item \textbf{CrisisMMD}~\cite{Alam_2018_CrisisMMD}. This real-world disaster response dataset includes around 35,000 noisy social media postings with paired images and text, each annotated for seven catastrophe categories and four severity levels. Its realistic user-generated content with natural noise and implicit modality correlations provides a rigorous testbed for zero-shot adaptation, with good performance indicating practical relevance in real-world crisis scenarios. 

\item \textbf{ScienceQA}~\citep{Lu_2022_ScienceQA}. This dataset contains 21,208 multimodal science questions combining textual and visual contexts, with 48.7\% of instances containing images. Questions span physics, chemistry, and biology domains, requiring cross-modal reasoning between textual concepts and visual diagrams. Additionally, ScienceQA offers image captions to aid text-only LLMs in reasoning, allowing a comparison of unimodal approaches.
\end{itemize}

\noindent\textbf{Task Formulation.}
For CrisisMMD, we define three multimodal classification tasks\footnote{This setting references the repository \href{https://github.com/PaulCCCCCCH/Multimodal-Categorization-of-Crisis-Events-in-Social-Media}{GitHub} and Abavisani \etal\cite{Abavisani_2020_CrisisMMD}}: (1) binary information relevance filtering, (2) fine-grained humanitarian category recognition, and (3) a consolidated taxonomy with merged categories to reduce label complexity. We omit the unimodal damage assessment to focus on multimodal aspects. For ScienceQA, we follow the original evaluation using multiple metrics: question types, contextual modalities, and educational stages. Performance is assessed through accuracy percentage across these categories.

\noindent\textbf{Baselines.} We conduct a comprehensive evaluation of text-only LLMs, multimodal VLMs, and KGs that enhance LLMs in multimodal reasoning. 
\begin{itemize}[noitemsep,nolistsep]
    \item For \textbf{CrisisMMD}, we compare text-only LLMs using few-shot prompting (LLaMA-2~\cite{Touvron_2023_Llama2}, GPT-4~\cite{Achiam_2023_Gpt4}, DeepSeek-R1~\cite{Guo_2025_Deepseekr1}, Qwen-2.5~\cite{Yang_2024_Qwen2}) and multimodal VLMs (CLIP~\cite{Radford_2021_Learning}, LLaVA~\cite{Liu_2023_LLaVa}, GPT-4o~\cite{Hurst_2024_Gpt4o}, Qwen2-VL~\cite{Wang_2024_QWEN2VL}, BLIP-2~\cite{Li_2023_BLIP2}).
    
    \item For \textbf{ScienceQA}, we compare models for general domains in zero/few-shot settings, including text-only LLMs (GPT Model~\cite{Lu_2022_ScienceQA}, CoT~\cite{Lu_2022_ScienceQA}, DDCoT~\cite{Zheng_2023_Advances}), multimodal VLMs (LG-VQA~\cite{Ghosal_2023_Language}, LaVIN~\cite{Luo_2023_LaVIN}, BLIP-2, CCOT~\cite{Mitra_2024_CCOT}, GraphVis~\cite{Deng_2024_GraphVis}) and Tool-LLM Chameleon~\cite{Lu_2023_Chameleon}. These models are not specifically fine-tuned for scientific tasks, ensuring a fair evaluation of generalization capabilities.
    
    \item We further compare the multimodal reasoning performance of LLMs assisted by KGs, evaluating text-based KGs built with LightRAG~\cite{Guo_2024_LightRAG}, and pre-constructed MMKGs such as Visual Genome~\cite{Krishna_2017_VisualGenome} and Mmkg~\cite{Liu_2019_MMKG}.
\end{itemize}

\noindent\textbf{Implementation.} For MMKG construction, we design a chain of VLMs including BLIP-2, LLaVA, and Qwen2-VL, with the CLIP-ViT-L/14 for pruning. Stronger or additional VLMs could be employed to enhance performance if more computational resources are available. We use the entire training set as the knowledge base and construct MMKGs from the extracted descriptions based on the LightRAG framework. In comparative experiments, the LightRAG method we evaluate utilizes only textual data, while VaLiK employs two configurations: (1) fully image-generated text descriptions (Image-only), and (2) original text combined with image-generated text (Text-Image). Dynamic window partitioning based on sentence length ensures syntactically coherent pruning results. Similarity thresholds are set to $\tau=0.25$ for CrisisMMD and $\tau=0.20$ for ScienceQA based on empirical evaluations to balance precision and recall. See Appendix E for selection details. We construct the graph using DeepSeek-R1-70B and implement LightRAG's hybrid retrieval approach with Qwen2.5-7B. For graph construction and multimodal reasoning, we utilize 1×NVIDIA A100-80GB GPUs. Task-specific prompts are designed to assist LLMs in multimodal reasoning evaluation.

\subsection{Main Results}

\begin{table*}[t!]
	\centering
	\resizebox{\textwidth}{!}{
		\begin{tabular}{lcccccccccc}
			\toprule
			\multirow{2}{*}{\textbf{Method}} &\multirow{2}{*}{\textbf{\#T-Param}} & \multicolumn{3}{c}{\textbf{Subject}} & \multicolumn{3}{c}{\textbf{Context Modality}} & \multicolumn{2}{c}{\textbf{Grade}} &  \multirow{2}{*}{\textbf{Average}} \\
			&& \textbf{NAT}   & \textbf{SOC}   & \textbf{LAN}   & \textbf{TXT}   & \textbf{IMG}   & \textbf{NO}    & \textbf{G1-6}  & \textbf{G7-12} &    \\ 
			\midrule
			Human~\cite{Lu_2022_ScienceQA} &-&  90.23 & 84.97 & 87.48 & 89.60 & 87.50 & 88.10 & 91.59 & 82.42 & 88.40 \\ \midrule
			GPT-4~\cite{Liu_2023_LLaVa} &-& 84.06 & 73.45 & 87.36 & 81.87 & 70.75 & 90.73 & 84.69 & 79.10 & 82.69 \\
                CoT (GPT-3)~\cite{Lu_2022_ScienceQA} &173B&  75.44 & 70.87 & 78.09 & 74.68 & 67.43 & 79.93 & 78.23 & 69.68 & 75.17 \\
			CoT (UnifiedQA)~\cite{Lu_2022_ScienceQA} &223M& {71.00} &  \underline{76.04} &  {78.91} &  {66.42} &  {66.53} & {81.81} &  {77.06} & 68.82 &  {74.11} \\
			CoT (GPT-4)~\cite{Lu_2023_Chameleon} &1T+& \underline{85.48} &  {72.44} &  \textbf{\underline{90.27}} &  {82.65} &  {71.49} & \textbf{92.89} &  \textbf{86.66} & 79.04 &  {83.99} \\
			DDCoT~\cite{Zheng_2023_Advances} &175B & 80.15 & \textbf{76.72} & 82.82 & 78.89 & 72.53 & 85.02 & 82.86 & 75.21 & 80.15 \\
                Chameleon (ChatGPT)~\cite{Lu_2023_Chameleon} &175B+& 81.62 &  70.64 &  84.00 &  79.77 &  70.80 & 86.62 &  81.86 & 76.53 &  79.93 \\  \midrule
                LG-VQA (BLIP-2)~\cite{Ghosal_2023_Language} &- & - &  -  &  -  & - &  - &  - &  - &  - & \textbf{86.32} \\
			LaVIN-13B~\cite{Yang_2023_Mm} &- & -	&- &-&	-&-	&- &	-& -& 77.54\\
                BLIP-2~\cite{Yang_2023_Mm} &- & -	&- &-&	-&-	&- &	-& -& 74.17\\
                CCOT &7B & -	&- &-&	-&-	&- &	-& -& 76.84\\
                GraphVis~\cite{Deng_2024_GraphVis} &7B & -	&- &-&	-&-	&- &	-& -& 73.18\\
			\midrule
			Qwen2.5-7B & 7B & 76.20 &  67.83  &  77.27  & 74.49 &  65.79 &  79.02 &  77.72 &  69.35 & 74.72 \\
                Qwen2.5-72B & 72B & 79.64 &  67.10  &  84.90  & 77.56 &  65.00 &  87.93 &  80.25 &  74.85 & 78.37 \\
                Qwen2.5-7B (Mmkg)~\cite{Liu_2019_MMKG} & 7B & 73.98 &  66.37  &  78.18  & 71.65 &  64.30 &  79.65 &  76.51 &  68.03 & 73.47 \\
                Qwen2.5-7B (Visual Genome)~\cite{Krishna_2017_VisualGenome}& 7B & 76.78 &  67.04  &  78.09  & 74.05 &  66.19 &  79.72 &  78.08 &  69.68 & 75.08 \\
			Qwen2.5-7B (VaLiK Text-only)& 7B & 84.54 &  74.24  &  86.91  & 82.74 &  72.53 &  90.03 &  84.51 &  80.28 & 82.98 \\
                Qwen2.5-7B (VaLiK Image-only)& 7B & 79.14 &  71.54  &  79.27  & 77.16 &  69.72 &  83.14 &  80.65 &  73.96 & 78.88 \\
                Qwen2.5-7B (VaLiK Text-Image)& 7B & 84.15 &  75.14 &  87.64 & \underline{82.99} &  \underline{73.18} &  89.69 &  84.40 &  \underline{80.95} & 83.16 \\
                Qwen2.5-72B (VaLiK Text-Image)& 72B & \textbf{85.61} &  75.93 &  \textbf{\underline{90.27}} & \textbf{84.40} &  \textbf{74.17} &  \underline{92.33} &  \underline{85.79} &  \textbf{82.98} & \underline{84.77} \\
			\bottomrule
		\end{tabular}
	}
	\caption{Performance comparison (\%) on ScienceQA benchmark. \#T-Params denotes trainable parameters. Categories: NAT (natural science), SOC (social science), LAN (language), TXT (text context), IMG-Cap (image caption), NO (no context), G1-6 (grades 1-6), G7-12 (grades 7-12).
    Method groups: (1) Human performance baseline, (2) Zero/Few-shot text-only LLMs, (3) Zero/Few-shot Multimodal VLMs, (4) LLMs enhanced with knowledge graphs for multimodal reasoning.
	}
		\label{tab:main}
\end{table*}

\textbf{Multimodal Classification Tasks.}
We conduct multimodal classification experiments on the CrisisMMD dataset, evaluating both text-only LLMs and multimodal VLMs. Detailed comparative results are provided in Tables~\ref{tab:text_kg} and~\ref{tab:multi_kg}.
For text-only LLMs, we adopt Qwen2.5-7B as the foundational reasoning model. Remarkably, the VaLiK-enhanced version achieves state-of-the-art (SOTA) performance matching that of the native Qwen2.5-72B model.
The image-only KG constructed through VaLiK demonstrates an average accuracy improvement of 4.41\% across tasks, with the text-image variant attaining a 4.90\% enhancement. These improvements significantly surpass the 1.22\% gain obtained by LightRAG using textual KG.
We further validate VaLiK's cross-scale applicability through evaluations on Qwen2.5-32B and 72B architectures, observing consistent 2.0\%–2.5\% improvements.
While not as significant as the 7B model's benefits, this shows that models that have substantial prior knowledge benefit less from external knowledge augmentation

Unlike text-only LLMs that depend on MMKGs for visual understanding, VLMs primarily benefit from KGs integration through outdated knowledge refreshment. Due to the inherent availability of visual features during inference, VaLiK's performance gains for VLMs remain constrained compared to text-only counterparts.
We separately applied VaLiK enhancement to Qwen2-VL-72B-Instruct and LLaVA-34B, obtaining distinct improvements: LLaVA-34B achieves accuracy gains of 2.41\% (image-only KG) and 3.59\% (text-image KG), while Qwen2-VL-72B-Instruct shows 1.77\% and 2.23\% improvements respectively under identical configurations. These experimental findings collectively demonstrate that VaLiK effectively extracts valuable signals from the training corpus and enables dynamic knowledge injection into VLMs during inference, thereby substantially alleviating hallucination phenomena. The differential improvements between Qwen2-VL-72B-Instruct and LLaVA-34B further validate the framework's adaptability across model architectures.

Additionally, we analyze the results of LLMs without KG enhancement in the tables, which generally follow the scaling law~\cite{Kaplan_2020_Scaling}. However, DeepSeek-R1 shows anomalous behavior. Through testing, we find that its reasoning process may introduce complex information that interferes with its judgment. Furthermore, empirical results show that most baseline models achieve suboptimal performance without fine-tuning. In contrast, VaLiK's automated MMKG construction framework requires no task-specific adaptation yet delivers consistent improvements.

\begin{table*}[t!]
\centering
\resizebox{\textwidth}{!}{
\begin{tabular}{llcccccccccc}
\toprule
\multirow{2}{*}{\textbf{Type}} & \multirow{2}{*}{\textbf{Method}} & \multirow{2}{*}{\textbf{\#T-Param}} & \multicolumn{3}{c}{\textbf{Subject}} & \multicolumn{3}{c}{\textbf{Context Modality}} & \multicolumn{2}{c}{\textbf{Grade}} & \multirow{2}{*}{\textbf{Average}} \\
& & & \textbf{NAT} & \textbf{SOC} & \textbf{LAN} & \textbf{TXT} & \textbf{IMG} & \textbf{NO} & \textbf{G1-6} & \textbf{G7-12} & \\
\midrule
\multirow{3}{*}{Image-Only} 
& Qwen2.5-7B & 7B & 76.20 & 67.83 & 77.27 & 74.49 & 65.79 & 79.02 & 77.72 & 69.35 & 74.72 \\
& + CVs & 7B & 80.06\,(\scriptsize{$\uparrow$3.86}) & 70.30\,(\scriptsize{$\uparrow$2.47}) & 80.55\,(\scriptsize{$\uparrow$3.28}) & 78.05\,(\scriptsize{$\uparrow$3.56}) & 68.43\,(\scriptsize{$\uparrow$2.64}) & 83.76\,(\scriptsize{$\uparrow$4.74}) & 81.17\,(\scriptsize{$\uparrow$3.45}) & 72.71\,(\scriptsize{$\uparrow$3.36}) & 78.14\,(\scriptsize{$\uparrow$3.42}) \\
& + SV & 7B & 79.14\,(\scriptsize{$\downarrow$0.92}) & 71.54\,(\scriptsize{$\uparrow$1.24}) & 79.27\,(\scriptsize{$\downarrow$1.28}) & 77.16\,(\scriptsize{$\downarrow$0.89}) & 69.72\,(\scriptsize{$\uparrow$1.29}) & 83.14\,(\scriptsize{$\downarrow$0.62}) & 80.65\,(\scriptsize{$\downarrow$0.52}) & 73.96\,(\scriptsize{$\uparrow$1.25}) & 78.88\,(\scriptsize{$\uparrow$0.74}) \\

\cmidrule(lr){1-12}
\multirow{3}{*}{Text-Image}
& Qwen2.5-7B & 7B & 76.20 & 67.83 & 77.27 & 74.49 & 65.79 & 79.02 & 77.72 & 69.35 & 74.72 \\
& + CVs & 7B & 81.88\,(\scriptsize{$\uparrow$5.68}) & 73.00\,(\scriptsize{$\uparrow$5.17}) & 84.00\,(\scriptsize{$\uparrow$6.73}) & 80.55\,(\scriptsize{$\uparrow$6.06}) & 70.05\,(\scriptsize{$\uparrow$4.26}) & 87.11\,(\scriptsize{$\uparrow$8.09}) & 82.01\,(\scriptsize{$\uparrow$4.29}) & 77.98\,(\scriptsize{$\uparrow$8.63}) & 80.57\,(\scriptsize{$\uparrow$5.85}) \\
& + SV & 7B & 84.15\,(\scriptsize{$\uparrow$2.27}) & 75.14\,(\scriptsize{$\uparrow$2.14}) & 87.64\,(\scriptsize{$\uparrow$3.64}) & 82.99\,(\scriptsize{$\uparrow$2.44}) & 73.18\,(\scriptsize{$\uparrow$3.13}) & 89.69\,(\scriptsize{$\uparrow$2.58}) & 84.40\,(\scriptsize{$\uparrow$2.39}) & 80.95\,(\scriptsize{$\uparrow$2.97}) & 83.16\,(\scriptsize{$\uparrow$2.59}) \\
\bottomrule
\end{tabular}}
\vspace{6pt}
\caption[ScienceQA Ablation Study]{Ablation study on ScienceQA benchmark (CVs: CoE-based Vision-Language Models; SV: Similarly Verification). Performance metrics include: NAT (natural science), SOC (social science), LAN (language), TXT (text context), IMG (image context), NO (no context), G1-6 (grades 1-6), G7-12 (grades 7-12).}
\label{tab:scienceqa_ablation}
\vspace{-10pt}
\end{table*}

\begin{table}[t]
\centering
\scalebox{0.70}{
\begin{tabular}{llccc}
\toprule
\multicolumn{1}{c}{\textbf{Type}} & \textbf{Method} & \textbf{Task 1 (\%)} & \textbf{Task 2 (\%)} & \textbf{Task 2-Merged (\%)} \\
\midrule
\multirow{3}{*}{Image-Only} 
& Qwen2.5-7B & 65.04 & 44.52 & 45.33 \\
& + CVs & 68.11\,(\scriptsize{$\uparrow$3.07}) & 47.00\,(\scriptsize{$\uparrow$2.48}) & 46.95\,(\scriptsize{$\uparrow$1.62}) \\
& + SV & 69.52\,(\scriptsize{$\uparrow$1.41}) & 49.54\,(\scriptsize{$\uparrow$2.54}) & 49.07\,(\scriptsize{$\uparrow$2.12}) \\[2pt]

\cmidrule(r){1-5}
\multirow{3}{*}{Text-Image}
& Qwen2.5-7B & 65.04 & 44.52 & 45.33 \\
& + CVs & 68.43\,(\scriptsize{$\uparrow$3.39}) & 48.61\,(\scriptsize{$\uparrow$4.09}) & 48.97\,(\scriptsize{$\uparrow$3.64}) \\
& + SV & 68.90\,(\scriptsize{$\uparrow$0.47}) & 50.02\,(\scriptsize{$\uparrow$1.41}) & 50.69\,(\scriptsize{$\uparrow$1.72}) \\
\bottomrule
\end{tabular}}
\vspace{6pt}
\caption{Ablation study on CrisisMMD with Qwen2.5-7B.}
\label{tab:crisismmd_ablation}
\vspace{-10pt}
\end{table}

\textbf{Multimodal Question Answering Tasks.} We evaluated multimodal QA performance on the ScienceQA benchmark with Qwen2.5-7B and Qwen2.5-72B as base architectures, augmented by four knowledge sources: Mmkg, Visual Genome, text-only LightRAG and VaLiK. Compared to existing zero-shot/few-shot LLMs that not specifically optimized for scientific QA, our VaLiK-enhanced Qwen2.5-72B achieved SOTA performance on 62.5\% of subtasks, demonstrating particular strengths in multimodal reasoning scenarios requiring cross-modal alignment with an average accuracy gain of 6.4\% over baseline models.

Our study identifies a fundamental imbalance between textual and visual knowledge representations in ScienceQA.
Text-only KGs (14k entities, 18k relations) exhibit 8× denser structured knowledge than image-only counterparts (3k concepts, 1k relations), explaining visual modality underperformance. Despite this gap, vision-KG-augmented Qwen2.5-7B still attains 4.16\% accuracy gains over its non-enhanced version.
Notably, our MMKG requires only 489MB storage for complete storage, while the scene graph component\footnote{\href{https://homes.cs.washington.edu/~ranjay/visualgenome/api.html}{Visual Genome}} of Visual Genome alone occupies 739MB. This lightweight construction enables effective reasoning using only textual KG descriptions without raw images in resource-constrained scenarios.

\subsection{Ablation Study}

Our ablation studies on \textbf{CrisisMMD} and \textbf{ScienceQA} demonstrate the specific roles of \textbf{VaLiK}'s components. As shown in Table~\ref{tab:scienceqa_ablation} and Table~\ref{tab:crisismmd_ablation}, the \textbf{CVs} (CoE-based VLM) module improves accuracy across all settings, with average gains of +3.05\% on CrisisMMD and +4.63\% on ScienceQA tasks, validating visual descriptions enhance reasoning. However, the \textbf{SV} (Similarly Verification) module exhibits dual effects: it significantly improves CrisisMMD metrics by pruning redundant textual descriptions, yet slightly degrades ScienceQA's image-only natural science reasoning. We hypothesize this discrepancy arises from dataset characteristics: CrisisMMD's generated captions contain substantially more redundant content, whereas ScienceQA's simpler visual scenes yield shorter descriptions. Pruning these shorter descriptions risks over-removal of critical semantics. Furthermore, different types of KGs influence the effectiveness of the components: CVs achieve greater gains in CrisisMMD's text-image fusion as original text provides complementary context, while SV shows reduced effectiveness, likely due to occasional over-pruning of cross-modal linkages. Nevertheless, both modules collectively enhance performance across configurations, demonstrating their synergistic yet context-sensitive nature.

\subsection{Further Analysis}
\label{subsec:analysis}

\begin{figure}[t]
	\centering
	\includegraphics[width=1.0\columnwidth]{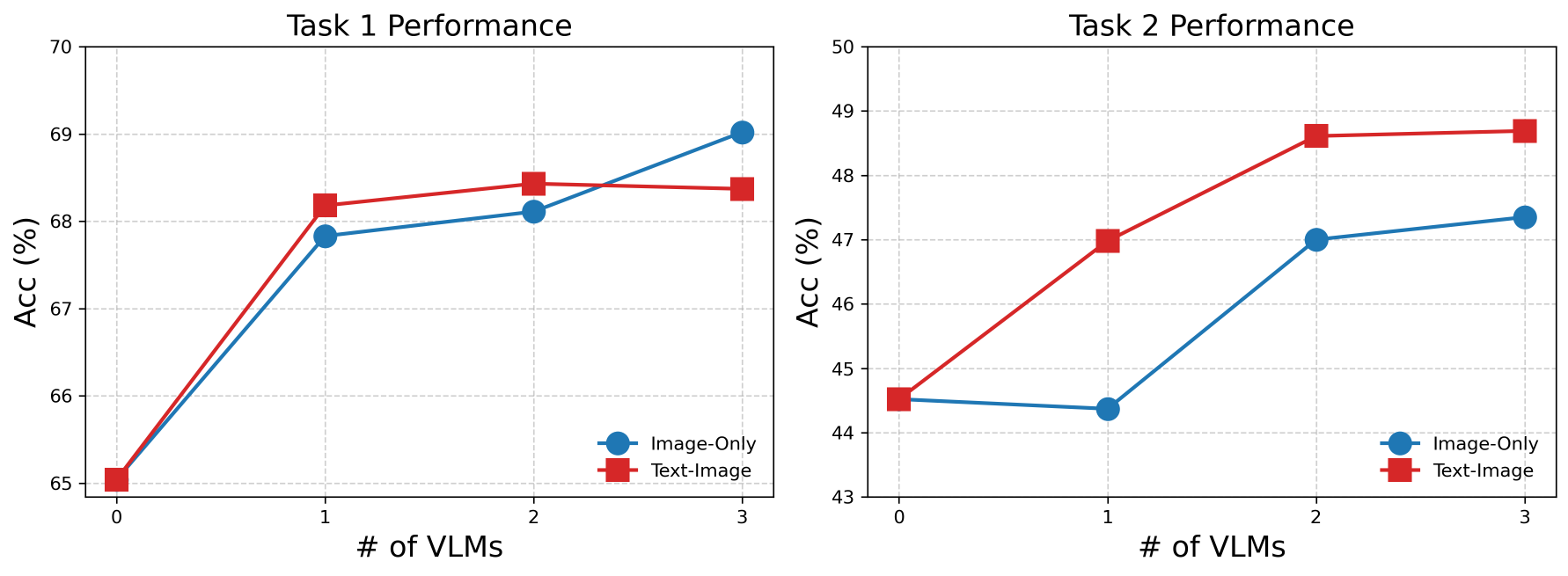}
	\caption{Impact analysis of VLM quantity on CrisisMMD.}
        \label{fig:chart}
    \vspace{-5mm}
\end{figure}

\textbf{Impact of VLM Quantity and Types.} We evaluate the impact of varying quantities and types of VLMs on the CVs module. Our experiments reveal that Qwen2-VL generates the most visual descriptions, followed by LLaVA, while BLIP-2 produces the fewest. However, BLIP-2 demonstrates superior capability in extracting critical information and identifying key entity relationships within images. We therefore adopt BLIP-2 as the primary model, with LLaVA or Qwen2-VL serving as secondary/tertiary components. Adding more VLMs yields diminishing returns, due to limited entities in current images, though we hypothesize their benefits would increase for complex visual scenes with richer semantic content. This phenomenon is empirically validated by our quantitative results in Figure~\ref{fig:chart}.

\textbf{Computational Costs.} Due to space limitations, we provide an overview of VaLiK's computational costs in Appendix F. Our method is significantly more cost-effective than manual annotation or LLM fine-tuning.

\section{Conclusion}
Multimodal reasoning in LLMs is constrained by incomplete knowledge and hallucination artifacts, limitations that persist because textual KGs cannot bridge visual-textual semantics due to their modality isolation. To bridge this gap, we propose VaLiK, a framework for constructing MMKGs through vision-language alignment, eliminating dependency on manual annotations while resolving visual-textual semantic inconsistencies. By integrating a cascade of pretrained VLMs and cross-modal verification, VaLiK converts images into structured knowledge while filtering noise. The resulting graphs enhance LLMs’ reasoning with minimal storage overhead. Experiments on multimodal reasoning benchmarks show SOTA performance. VaLiK’s modular design supports adaptability across domains, offering a scalable solution for autonomous knowledge synthesis. This work advances multimodal AI systems by enabling efficient integration of visual and textual data.

\section{Acknowledgments}
The research was supported by Shanghai Artificial Intelligence Laboratory, the National Key R\&D Program of China (Grant No. 2022ZD0160201) and the Science and Technology Commission of Shanghai Municipality (Grant No. 22DZ1100102).

{
    \small
    \bibliographystyle{ieeenat_fullname}
    \bibliography{main}
}
 
\appendix
\clearpage
\setcounter{page}{1}
\maketitlesupplementary

\section{Cross-Modal Reasoning Failures in Textual KGs}
\label{sec:rationale}

\begin{figure}[t]
	\centering
	\includegraphics[width=1.0\columnwidth]{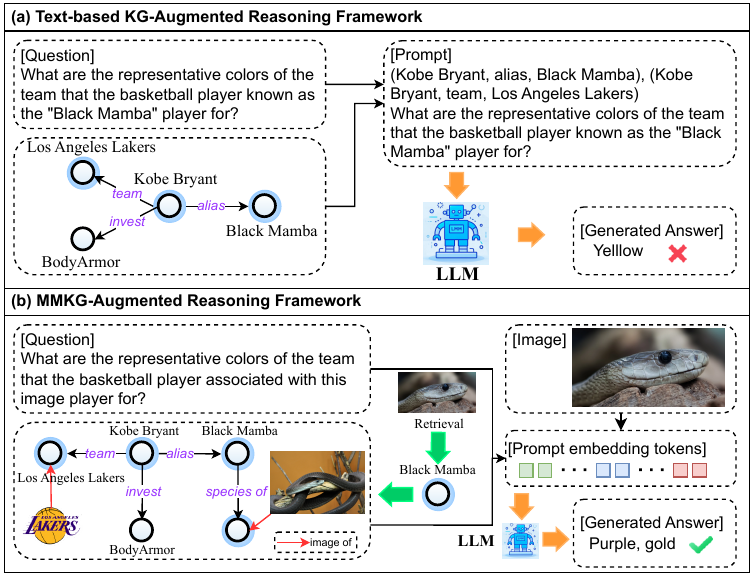}
	\caption{(a) The limited information contained in text-based KGs leads to inaccurate responses. (b) Leveraging MMKGs enables reasoning with enriched multimodal information to produce the correct answer. }
        \label{fig:unimodal_multimodal}
    \vspace{-4mm}
\end{figure}

Multimodal learning, by virtue of its capability to synergistically integrate heterogeneous data modalities, establishes a comprehensive knowledge acquisition paradigm that significantly enhances reasoning robustness \cite{Lee_2024_Multimodal}. This principle extends to Multimodal Knowledge Graphs (MMKGs), where the semantic symbiosis between visual and textual modalities addresses the critical limitation of modal isolation inherent in conventional text-based KGs. As empirically demonstrated in Figure~\ref{fig:unimodal_multimodal}, pure textual KGs often induce hallucinated or incomplete responses due to their inability to resolve visual-textual semantic ambiguities. For instance, when queried about fine-grained visual attributes (e.g., spatial relationships or object properties absent in textual metadata), LLMs grounded solely on textual KG triples frequently generate plausible but factually inconsistent answers, as they lack access to cross-modal referential grounding. In contrast, MMKGs bridge this gap through bidirectional visual-textual entity linking, enabling LLMs to retrieve and reason over fused evidence from both modalities. Our qualitative analysis of the case in Figure~\ref{fig:unimodal_multimodal} reveals that the multimodal reasoning path—leveraging both image-derived entities and textual relationships—is essential for deriving logically coherent and factually accurate conclusions.

\section{Case Studies on Manual Annotation Overheads}

\begin{figure}[t]
	\centering
	\includegraphics[width=1.0\columnwidth]{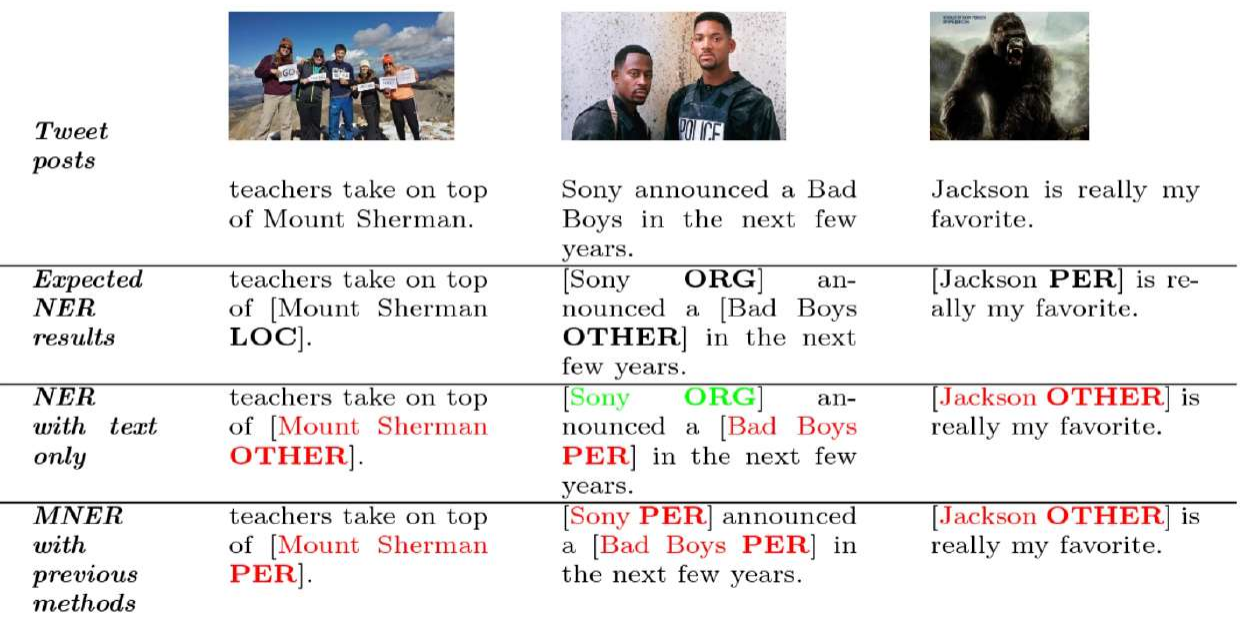}
	\caption{Three example social media posts with labelled named entities \cite{chen2021multimodal}.}
        \label{fig:example}
    \vspace{-4mm}
\end{figure}

\begin{table}[t]
\centering
\label{tab:coreference_stats}
\scalebox{0.95}{
\begin{tabular}{lrrr}
\toprule
\textbf{Type} & \textbf{\#Chains} & \textbf{Mentions/Chain} & \textbf{Boxes/Chain} \\ 
\midrule
people        & 59766 & 3.17 & 1.95 \\
clothing      & 42380 & 1.76 & 1.44 \\
body parts    & 12809 & 1.50 & 1.42 \\
animals       &  5086 & 3.63 & 1.44 \\
vehicles      &  5561 & 2.77 & 1.21 \\
instruments   &  1827 & 2.85 & 1.61 \\
scene         & 46919 & 2.03 & 0.62 \\
other         & 82098 & 1.94 & 1.04 \\
\midrule
total         & 244035 & 2.10 & 1.13 \\
\bottomrule
\end{tabular}
}
\caption{Coreference chain statistics of Flickr30K-Entity. The number of mentions per chain indicates how salient an entity is. The number of boxes per chain indicates how many distinct entities it refers to.}
\end{table}

The development of robust entity extraction models typically hinges on large-scale annotated corpora, yet the generalizability of these models remains intrinsically bounded by the semantic scope and granularity of their training datasets. Widely-adopted benchmarks such as Flickr30K-Entity \cite{Plummer_2015_Flicker30k} exemplify this constraint: while serving as de facto standards for evaluating visual-linguistic entity grounding, their construction necessitates labor-intensive manual annotations at scale. As illustrated in Figure~\ref{fig:example}, even high-quality annotations in such datasets often adopt a minimalist tagging paradigm—identifying only coarse-grained entities while neglecting fine-grained attributes and contextual relationships. This sparsity of semantic enrichment directly propagates to trained models, which consequently fail to capture the compositional semantics necessary for complex reasoning scenarios.

\section{Case Studies on Visual Specificity Deficits in VLM-Generated Captions}

\begin{figure}[t]
	\centering
	\includegraphics[width=1.0\columnwidth]{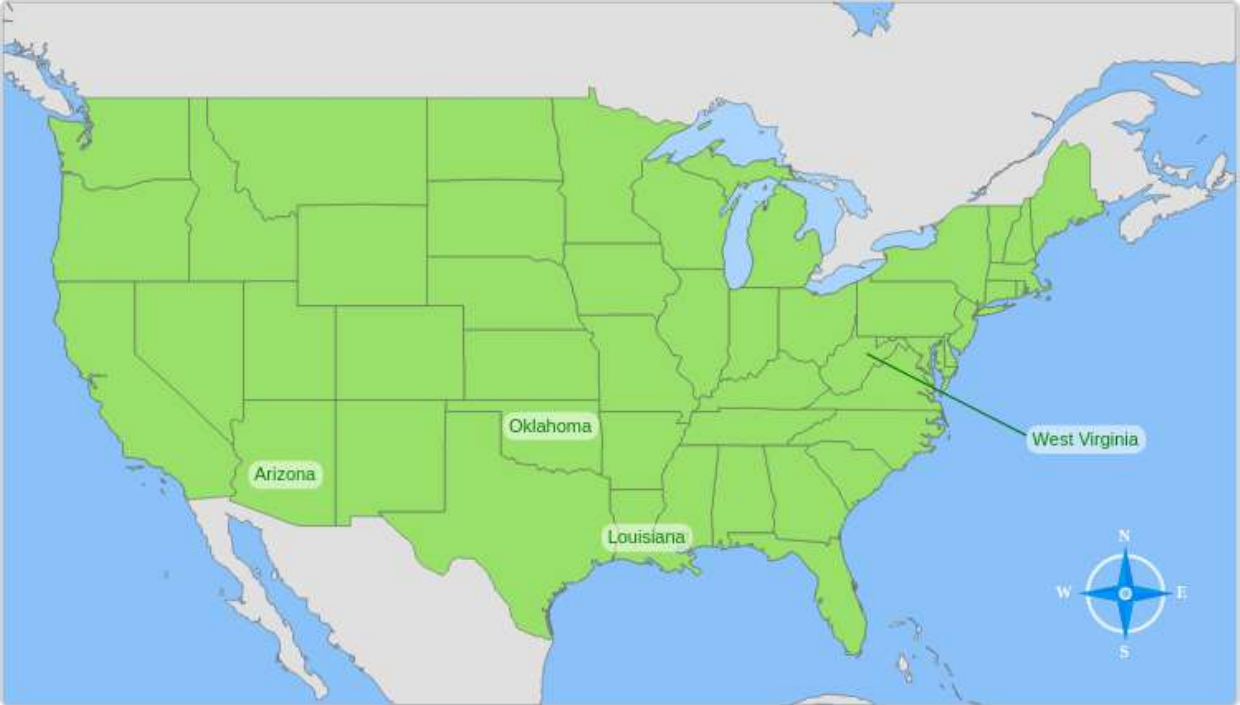}
	\caption{An example from the ScienceQA benchmark \cite{Lu_2022_ScienceQA}, illustrating multimodal question-answering scenarios that necessitate joint reasoning over textual prompts and visual evidence.}
        \label{fig:map}
    \vspace{-4mm}
\end{figure}

As exemplified in Figure~\ref{fig:map}, vision-language models like BLIP-2 \cite{Li_2023_BLIP2} tend to produce oversimplified textual descriptions that critically lack actionable visual-semantic signals. The VLM-generated caption \textbf{("A map of the united states with the location of the united states")} merely identifies coarse-grained scene semantics, failing to capture object-level attributes (color coding of regions), spatial relationships (border adjacency between Arizona and Mexico) and compositional context (compass orientation in lower-right corner). In contrast, human annotations \textit{("This is a map of the United States. The main part of the country is shown in green, with several states labeled. Arizona is in the southwestern part of the US, bordering Mexico. Oklahoma is in the central - southern region. Louisiana is located along the Gulf of Mexico in the southeastern part. West Virginia is in the eastern part of the country. There's also a compass in the bottom - right corner to show directions.")} demonstrate essential characteristics for multimodal reasoning.

\section{Retrieval Strategy in MMKG Construction}

We adopt retrieval strategies based on the framework provided by LightRAG~\cite{Guo_2024_LightRAG}, which supports multiple modes:
\begin{itemize}
    \item \textbf{local}: focuses on context-dependent information;
    \item \textbf{global}: utilizes global knowledge;
    \item \textbf{hybrid}: combines local and global retrieval methods;
    \item \textbf{naive}: performs basic search without advanced techniques;
    \item \textbf{mix}: integrates knowledge graph and vector retrieval;
\end{itemize}
In our implementation, we rely on the \textbf{hybrid} retrieval mode, which balances the precision of local cues with the breadth of global knowledge. This strategy improves the relevance and completeness of retrieved information, which is crucial for high-quality MMKG construction.

LightRAG is an excellent project that effectively supports automatic MMKG construction, and its retrieval design plays a central role in our framework. Specifically, LightRAG introduces keyword-guided text chunking to expand the retrievable context. By leveraging both high-level and low-level keywords in combination with chunk-level vector retrieval, it enables more comprehensive knowledge access.
In addition, the choice of the retrieval model is also important. Larger LLMs have slower retrieval speeds but better performance. In this experiment, we used Qwen2.5-7B for retrieval. We also tested the retrieval performance of 32B and 72B models, which showed a 1\%-5\% improvement in performance, but it also significantly increased the graph construction time. Therefore, we finally adopted a lightweight retrieval model.
The details of the entire LightRAG are shown in Algorithm ~\ref{alg:kg-generation}.

\begin{algorithm}[t]
    \caption{MMKG Generation}
    \label{alg:kg-generation}
    \begin{algorithmic}[1]
        \Require 
            \( \hat{S} \) (refined description), 
            \( T \) (external knowledge, optional)
        \Ensure 
            \( \mathcal{G} = (\mathcal{E}, \mathcal{R}) \) (knowledge graph)
        
        \State \( \mathcal{T} \gets \hat{S} \oplus T \)  \Comment{Concatenate \( \hat{S} \) and \( T \)}
        
        \State \( \mathcal{G} \gets \text{LightRAG}(\mathcal{T}) \)  \Comment{Generate graph via LightRAG}
        
        \State \( (\mathcal{E}, \mathcal{R}) \gets f_{\text{ERE}}(\mathcal{T}) \)  \Comment{Extract entities and relations}
        
        \State \Return \( \mathcal{G} = \{(h, r, t) \mid h,t \in \mathcal{E}, r \in \mathcal{R}\} \)
    \end{algorithmic}
    
\end{algorithm}

\section{\texorpdfstring{Selection of Sensitivity Threshold \(\tau\)}{Selection of Sensitivity Threshold tau}}

We select the sensitivity threshold \( \tau \) empirically based on performance on the validation set. In practice, \( \tau \) can be approximately determined by observing the token length distribution of captions: datasets with richer visual content and longer captions tend to benefit from a lower \( \tau \), while simpler datasets can tolerate a higher \( \tau \). This provides a practical way to adjust \( \tau \) without extensive tuning.

In addition, we notice a key pattern when analyzing the relevance scores across windows. Around certain values of \( \tau \), the scores tend to cluster tightly on both sides of the threshold. As a result, even a small change in \( \tau \) near these points can lead to a large change in the number of tokens being pruned. This indicates that the pruning process is especially sensitive around those points, and adjusting \( \tau \) even slightly may have a big impact on the final token budget.

\section{Construction Cost and Scalability}

Construction cost is a complex issue, which we analyze from the perspectives of time and hardware requirements. Time-wise, the main components are CoE and LightRAG. While using APIs can significantly speed up the process, offline deployment and inference are also feasible. For example, generating descriptions with Qwen2-VL-7B achieves around 60 tokens per second, processing one image every 4 seconds. Thus, processing 1k images takes approximately 1.21 hours. Constructing a KG with Qwen2.5-7B yields about 196k tokens per hour, leading to a total of 1.33 hours for 1k images. The intermediate pruning step, accelerated by CLIP’s fast processing speed, is negligible.
Overall, the cost is much lower than manual annotation or fine-tuning LLMs, making the method applicable to large-scale datasets. For resource-constrained users, deploying a lightweight VLM with CoE is comparable to or even more efficient than deploying a powerful VLM, further demonstrating the scalability of our approach.

\section{Discussion on VLM Usage and Design Flexibility}

Our observations on the number and type of VLMs used in CoE are consistent with the original conclusions drawn in the CoE paper~\cite{Xiao_2024_CoE}. Regardless of the specific VLM architecture, increasing the number of models \( N \) consistently improves performance up to a saturation point, after which further scaling yields diminishing returns.
Moreover, we find that convergence is achieved more quickly when using lower softmax temperatures or simpler datasets. These factors reduce the ambiguity in model disagreement, allowing consensus to form more rapidly among the ensemble.

Interestingly, our results also show that using a single, strong VLM can achieve performance comparable to a cascade of smaller, lightweight models. This suggests a practical trade-off between model strength and ensemble size—while ensembling helps in reaching consensus across diverse weak learners, a single high-capacity model may suffice in many scenarios, especially when computational resources are limited.

In the original CoE method, the outputs from all VLM experts are first aggregated together, and then a selection process determines which expert descriptions to use. To save time in constructing the MMKGs with LLMs, we instead adopted a sequential strategy where the output of one expert is used as the prompt input for the next.
We also evaluated the original aggregation and selection strategy on a smaller-scale dataset and found it to perform well, sometimes even surpassing the sequential approach. This confirms that CoE's original design of aggregating all experts' outputs before selecting which descriptions to use is effective and remains a strong baseline. However, correspondingly, using LLMs to construct MMKGs based on these aggregated descriptions requires significantly more time.

Additionally, while we apply pruning only at the final description step, pruning during intermediate steps may also yield good results depending on the dataset and task. There is no fixed rule for when or how to apply pruning, and our framework is designed to be flexible enough to accommodate different strategies.
We emphasize that both our CoE framework and the SV step are intended to be adaptable, allowing users to experiment freely and select the approach that best suits their needs.

There are various VLMs that can be used for pruning. Among them, we recommend CLIP due to its fast inference speed and pruning performance comparable to other VLMs. Given its efficiency and effectiveness, CLIP serves as a practical choice for pruning in many scenarios.

\end{document}